%% file: main.tex
\documentclass{article}

\PassOptionsToPackage{sort, numbers}{natbib}

\usepackage[final]{nips_2017}

\usepackage[utf8]{inputenc}
\usepackage[T1]{fontenc}
\usepackage{hyperref}
\usepackage{url}
\usepackage{booktabs}
\usepackage{amsfonts}
\usepackage{nicefrac}
\usepackage{microtype}
\usepackage{grffile}
\usepackage{amsmath}
\usepackage{graphicx}
\usepackage{verbatim}
\usepackage{setspace}
\usepackage{subcaption}
\usepackage{wrapfig}
\usepackage{xcolor}
\title{Visual Decoding of Targets During Visual Search From Human Eye Fixations\\}

\author{
	Hosnieh Sattar \\
	Max Planck Institute for Informatics \\
	Saarbr\"{u}cken, Germany \\
	\texttt{sattar@mpi-inf.mpg.de} \\
	\And
	Mario Fritz \\
	Max Planck Institute for Informatics \\
	Saarbr\"{u}cken, Germany \\
	\texttt{mfritz@mpi-inf.mpg.de} \\
	\AND	
	Andreas Bulling \\
	Max Planck Institute for Informatics \\
	Saarbr\"{u}cken, Germany \\
	\texttt{bulling@mpi-inf.mpg.de} \\
}

\begin{document}

\maketitle

\input{abstract.tex}

\input{Introduction.tex}

\input{Relatedworks.tex}

\input{Methods.tex}

\input{ExperimentsandResults.tex}

\newpage
\begin{spacing}{0.8}
	\small
	\bibliography{refs}
\end{spacing}

\end{document}

%% file: abstract.tex
\begin{abstract}
What does human gaze reveal about a users' intents and to which extend can these intents be inferred or even visualized?
Gaze was proposed as an implicit source of information to predict the target of visual search and, more recently, to predict the object class and attributes of the search target.
In this work, we go one step further and investigate the feasibility of combining recent advances in encoding human gaze information using deep convolutional neural networks with the power of generative image models to visually decode, i.e. create a visual representation of, the search target.
Such visual decoding is challenging for two reasons: 1) the search target only resides in the user's mind as a subjective visual pattern, and can most often not even be described verbally by the person, and 2) it is, as of yet, unclear if gaze fixations contain sufficient information for this task at all.
We show, for the first time, that visual representations of search targets can indeed be decoded only from human gaze fixations.
We propose to first encode fixations into a semantic representation and then decode this representation into an image.
We evaluate our method on a recent gaze dataset of 14 participants searching for clothing in image collages and validate the model's predictions using two human studies.
Our results show that $62\%$ (Chance level = 10\%) of the time users were able to select the categories of the decoded image right. In our second studies we show the importance of a local gaze encoding for decoding visual search targets of users.
\end{abstract}

%% file: Introduction.tex
\section{Introduction}

Predicting the visual search target of users is important for a range of applications, particularly human-computer interaction, given that it reveals information about users' search intents.
During visual search, gaze conveys rich information about the target a user has in mind. Recent development of eye tracking technology makes usage of gaze data in different task more affordable and is expected to become more ubiquitous in the future.
Using gaze data for recognizing the search target or intent can be considered very convenient for users as gaze is captured implicitly and users don't need to verbalized or give any implicit input to the machine for assistance. 

More broadly, we want to understand to what level of detail such  search targets and intends can be recovered.
Recently, several studies shown the possibility of visual search target prediction from gaze\cite{SattarBF16, sattar15_cvpr,borji2014eyes,zelinsky2013eye}. 
Pioneer works tried to predict the search target in the limited set of known targets from gaze data\cite{zelinsky2013eye,borji2014eyes}. More recent works, move smoothly from closed set of targets to an open-world setting \cite{sattar15_cvpr}. Most recently, \cite{SattarBF16} introduced the gaze pooling layer that can recognize categories and attributes of visual search target. They also removed the need of gaze to train a classifier to recognize the search target of users. This enables the usage of  gaze in conjuncture with state of the art deep learning recognition system to recognize users intent.  Yet, previous efforts on analyzing search targets and intents have only addressed a classification scenario -- hence a finite set of  discrete classes.

On the other hand,  cognitive neuro scientists have shown first success of visualizing images based on fMRI data \cite{Cowen201412,Nishimoto20111641}.
While we also want to access aspects of the mental state,  our task is fundamentally different in two aspects: (1) While they aim at decoding a specific image that was shown to a person, our goal is to decode visual search target. (2) While they are using fMRI data, we are are using gaze data. On a related note, we believe that gaze data is particular interesting to investigate, as it is practical and affordable to collect and use in many application scenarios of future interfaces \cite{sato2016sensing,zhang2015appearance}. 

As we are targeting decodings of categorical search targets, generating visualization is challenging due to strong intra-class variations.
However, recent advances in deep learning have led to a new generation of generative models for images. Recently, \cite{Yan2016}
generates images of objects from high-level descriptions. They transfer the high-level text informations to a set of attributes (e.g hair color: Brown, gender: female). These attributes are used later on to build an attribute-conditioned generative model.

Hence, we approach  decoding of search targets from gaze data by bringing together recent success in gaze encoding and categorization with state of the art category conditioned generative image models.

The main contributions of this work are:
\begin{itemize}
\item  First proof of concept that visual representations of search targets can be  decoded from human gaze data.
\item  We present a practical approach, as it respects the difficulties in collecting large human gaze datasets.  Encoder and decoders are trained from large image corpora and transfer between the two representation is facilitated by a semantic layer in between.
\item We show the importance of localized gaze information for improved search target reconstruction.
\end{itemize}

%% file: Relatedworks.tex
\section{Related Work}

Our work is informed by previous works on gaze-based search target prediction, as well as generative models for image synthesis. 

\paragraph{Predicting Search Targets From Gaze}

Several previous works focused on predicting the search targets of users from gaze behavior.
Zelinsky et al.\ predicted search targets from subjects' gaze patterns during a categorical search task~\cite{zelinsky2013eye}. 
In their experiments, participants were asked to find two categorical search targets among four visually similar distractors.
Borji et al.\ focused on predicting search targets from fixations~\cite{borji2014eyes}.
They used a binary pattern and 3-level luminance patterns as target and participants were asked to find targets out of a set of other patterns.
A compatibility measure was introduced by \cite{sattar15_cvpr} to predict the search targets of users in an open and closed-world setting.
In a follow-up work \cite{SattarBF16}, they introduced the gaze-pooling layer to integrate gaze data into CNN to predict the category and attributes of search targets.
In this work, we are inspired by their work as we also use a gaze-pooling layer as gaze encoder. Yet, our work is first to address reconstruction of the search target from fixation data, which is more challenging than prediction as it addresses a continuous output space. 

\paragraph{Image generation and multi-modal learning}

Due to recent advances of representation learning and convolution neural networks, image generation becomes possible.
Recently, generative adversarial networks (GANs)~\cite{NIPS2014_5423,reed2016generative,NIPS2015_5773,pix2pix2016,pathakCVPR16context}
were used to generate realistic and novel images.
GANs consists of two parts: a generator and discriminator. The discriminator is designed to discriminate between generated images and training data.
However, training GANS is a challenging task due to the min-max objective. 
A stochastic variational inference and learning algorithm was introduced by
\cite{kingma2013auto}. 
A lower bound estimator is achieved via re-parameterization of the variational lower bound. Consequently, standard stochastic gradient method can be used to optimize the estimator. 
However, the posterior distribution of latent variables is usually unknown.
Yang et al.\ introduced a general-optimization based approach that uses image generation models and latent priors for posterior inference~\cite{Yan2016}. They generated images conditioned on visual attributes.
In our work, we employ their idea of conditional generative models in the context of inferring search intends from gaze data. 

\paragraph{Visual Experience Reconstruction using fMRI}

Recent developments in functional magnetic resonance imaging (fMRI) make it possible for neuroscientist to generate links between brain activity and the visual world.
In a more advanced setting, Nishimoto et al.\ reconstructed natural movies from brain activity~\cite{Nishimoto20111641}.
They proposed a motion energy encoding methods to decode the fast visual information and BOLD signals in occipitotemporal visual cortex and fit the model separately to individual voxels.

In another work, Cowen et al.\ proposed to reconstruct human faces from evoked brain activity using multi-variant regression and PCA~\cite{Cowen201412}.
In their experiment they asked participants to look at an image and then tried to reconstruct this specific image from fMRI data.
All of the aboves task tried to reconstruct the {\it seen} images. In constrast, our approach decodes the {\it visual search target} of user's which only resides in user's mind. Also, we are not using fMRI but gaze data which is arguably more practical and affordable to collect and use.

%% file: Methods.tex
\section{Method}
\begin{figure*}[t]
\begin{center}
\includegraphics[width=1\linewidth]{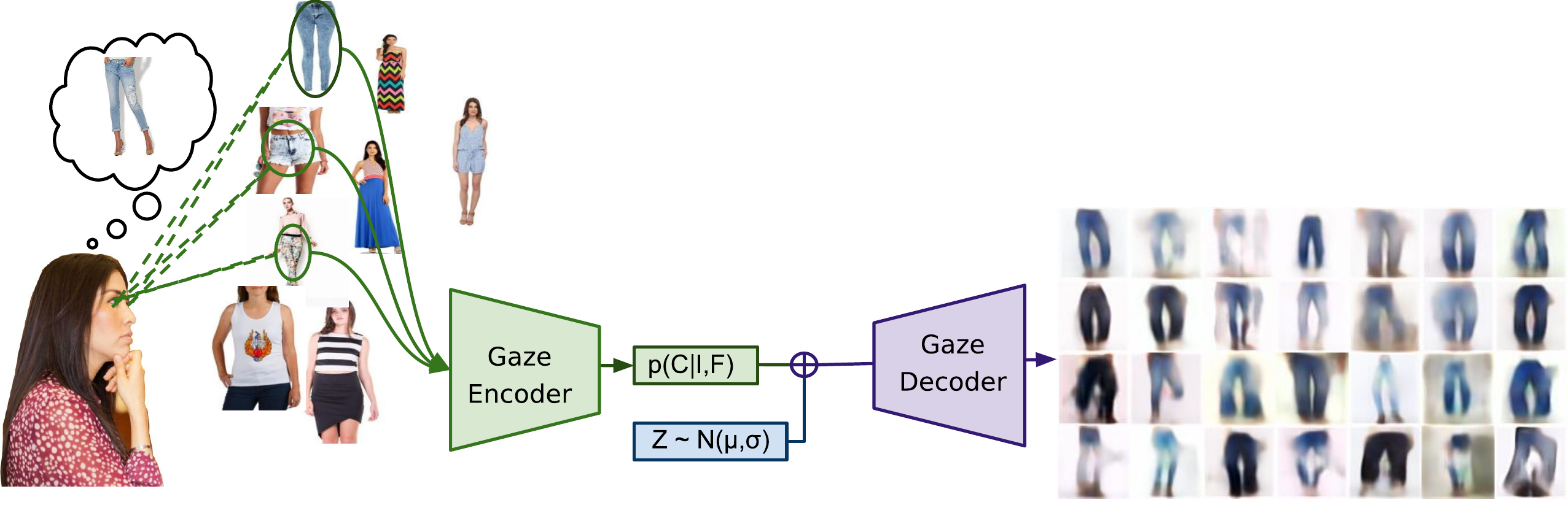}
\end{center}
   \caption{This image gives an overview of our approach. The user is searching for a category ``Jean'', the gaze data is recorded during the search task. We encode the gaze information into a semantic representation $p(C|I,F)$. The representation is used as condition over the learned latent space to decode the gaze into visualizations of the categorical search target.}
\label{fig:short}
\end{figure*}

In this work we address decoding of users visual search target from gaze. In order to achieve our goal, we build on recent success in gaze encoding and categorization as well as state of the art category conditioned generative image models. The gaze encoding is used to encode the raw gaze data into a semantic categorical space. The generative image model is conditioned on the encoded gaze data to decode visual search target of users as illustrated in \autoref{fig:short}. 

More formally, participant $P\in \mathbb P$ were searching for a target category $C \in \mathbb{C}$
in the collage image $I$. During the search task users performed fixations $F(I,C,P)={(x_i,y_i,t_i),i=1,...,N}$, where each fixation is a triple of positions $x_i,y_i$ in screen coordinates and fixation duration $t_i$.
our goal is to sample the visual search target $ST$ of the target category $c$ from
\begin{align}
P(\textit{ST}|\textit{F}(I,C,P)) = \sum_c P(\textit{ST}|c)P(c|\textit{F}(I,C,P)),
\end{align}
\noindent
where $P(c|\textit{F}(I,C,P))$ corresponds to the encoding of the gaze data into a semantic space of $c$ and  $P(\textit{ST}|c)$ a decoding from that semantic space to the visual  search target.
In the following, we explain the gaze encoder and search target  decoder.

\subsection{Semantic Gaze Encoder}
following after\cite{SattarBF16}, we represent fixations with a Fixation Density Map (FDM):
\begin{align}
\textit{FDM}(F) =  \sum_{f\in F} \textit{FDM}(f)
\end{align}
where each fixation $f$ is represented by a Gaussian spatial distribution of fixed variance at the location of the fixated point \textit{FDM}(f). The FDM are then combined with visual features F(I), obtain from a GAP Deep Learning Architecture \cite{zhou2016cvpr} in a Gaze Pooling Layer. The integration is done via element-wise multiplication of FDM and F(I):

\begin{align}
\textit{GAP}_\text{GWFM}(I,F)=\sum_{x,y}F(I)\times \textit{FDM}(F)
\end{align}

To get a final class prediction the weighted feature maps are averaged and fed into a fully connected and \textit{softmax} layer: 

\begin{align}\label{Encode}
p(C | I,F)=\textit{softmax}(W \;\textit{GAP}_\text{GWFM}(I,G) +b),
\end{align}

$W$ are learned weights and $b$ is the bias. 
We use those class posterior as gaze encoding and condition the generation model on them. We hypothesis that this encoding convey rich information on user's visual search target. 

\subsection{Visual Search Target Decoder}
In order to  sample visual search targets of users, we employ a generative image model that we condition on  the class posteriors predicted via gaze pooling layer and a latent random variable $z$.
\subsection*{Category-Conditioned Image Generation model}
Conditional Variational Auto-Encoder (CVAE) \cite{Yan2016} has been shown successfully sample images based on attribute classes. Here, we build on this recent success and train the model for generation of images of different clothing categories. Given the category vector $y\in \mathbb{R}^{d_y}$ and the latent variable $z \in \mathbb{R}^{d_z}$, our goal is to build a generative model $p_\theta (x|y,z)$, which generates image $x \in \mathbb{R}^{d_x}$.
The generated image is conditioned on the categorical information and the latent variable.
In conditional variational auto-encoder, the auxiliary distribution $q_\phi(z|x,y)$ is introduced to approximate the true posterior $p_\theta(z|x,y)$. The goal of learning process is to find the best parameter $\theta$ which maximizes the lower bound of the log-likelihood $log~p_\theta(x|y).$
Hence the conditional log-likelihood is 

\begin{align}
 log~p_\theta(x|y) = \textit{KL}(q_\phi(z|x,y)||p_\theta(z|x,y))+\mathcal{L}_{CVAE}(x,y,;\theta,\phi),
\end{align}

where the variational lower bound

\begin{align}
\mathcal{L}_{\text{\it CVAE}}(x,y,;\theta,\phi) = -\textit{KL}(q_\phi(z|x,y)||p_\theta(z|x,y))+\mathbb{E}_{q_\phi(z|x,y)}[log~p_\theta(x|y,z)]
\end{align}

is maximized for learning the model parameter. 
\subsection{Pruning Strategies}\label{sec:implementation}
We describe the details of our method and the design choices that make our method more robust to noise. After all,  human gaze data is noisy and needs aggregation over multiple observations in order to extract information. In the following, we describe how we aggregate information across multiple stimuli as well as  prune the  representation of the gaze encoder.

\paragraph{Gaze Encoding over Multiple Stimuli} Typical dataset for human gaze contains, data over multiple stimuli. For the dataset that we are going to use in the next section, such stimuli correspond to multiple collages that are shown to the participants. We compute weighted average of posteriors of fixated images with fixation durations to get a prediction over a collage. Moreover, we also average the resulted posteriors per collage to get one final posterior over the set of collages for each categories. 

\paragraph{Coping with Noise in Gaze Encoder Prediction}
Despite the aggregation as described above, inferred quantities from gaze are typically still noisy. This can also be seen from the prior study using the gaze encoder \cite{SattarBF16}. In order to cope with this challenge, we try different pruning strategies to suppress weak activations in the semantic representation $c$.

Specifically, we tried four scenarios to decode visual search target from gaze. In the first case, we used the plain posterior as conditioned vector. In the remaining cases, we used only the top 1 to top 3 highest activated classes in the posterior as condition vector for CVAE. All other probabilities are set to zero and the posterior is renormalized afterwards.

%% file: ExperimentsandResults.tex
\section{Experiments}
We evaluate our approach on  an existing dataset of human gaze. Besides qualitative results, we show two user studies. One measures the success in reconstructing meaningful visual representations of visual search targets and the second highlight the importance of  a gaze encoder that respects localized information. We start by describing the dataset we use throughout our experimental evaluation.

\subsection{Dataset}
We evaluate our model on the gaze dataset from \cite{SattarBF16}. Their data set contains gaze data of 14 participants searching for 10 different clothing categories and attributes. They used Deep-fashion\cite{zhou2016cvpr} dataset to train their proposed network and synthesizes their collages. Consequently, we also used Deep-fashion data set to train the CVAE for image generation. 
In addition, we train a classifier (VGGNet-16-GAP) as explained in \cite{SattarBF16} to get class posterior from gaze data. Both networks are trained over top 10 categories of Deep-fashion. We use the same  train, test and validation split as proposed in the deep-fashion dataset.

\subsection{Qualitative Results of Search Target Decoding}
\begin{figure*}
\begin{center}
\includegraphics[width=1\linewidth]{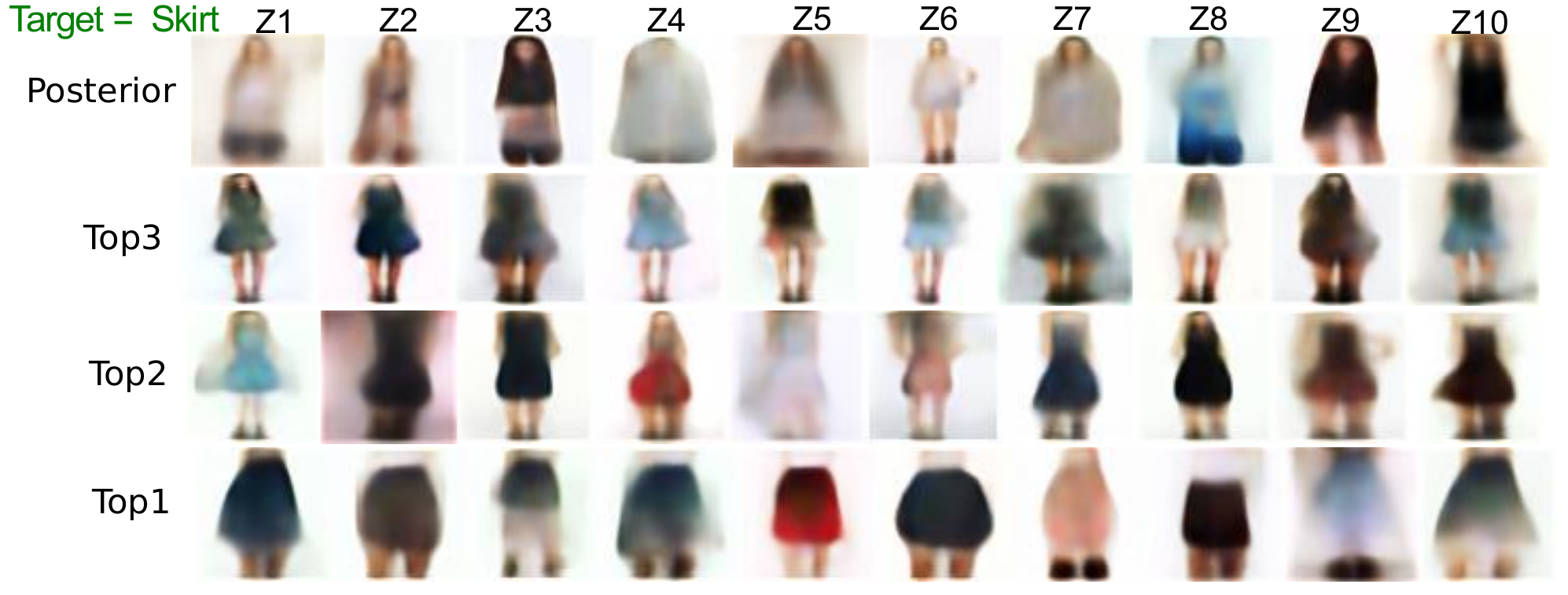}
\end{center}
   \caption{Using all posteriors gives images that contain several categories. Using only top3 to top1 posterior, gives images which contain the intended categories. As we move from posteriors to top1 the decoded image is more localized and contains fewer classes. Top3 images have full body part, as we move to top 1, can see only lower body part that contains a skirt.}
\label{fig:ave}
\end{figure*}

\begin{figure*}
\begin{center}
\includegraphics[width=0.9\linewidth]{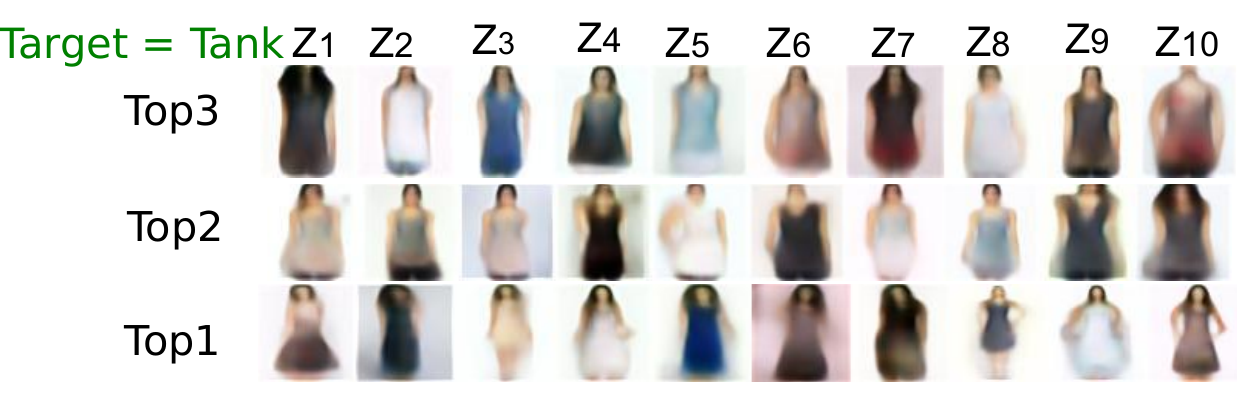}
\end{center}
   \caption{Top3 and top2 were able to capture the right category, the decoded images contain the target ``Tank''. However, due to wrong prediction for top1 resulted decoding looks like a ``Dress''.}
\label{fig:short}
\end{figure*}

\begin{figure*}
\begin{center}
\includegraphics[width=1\linewidth]{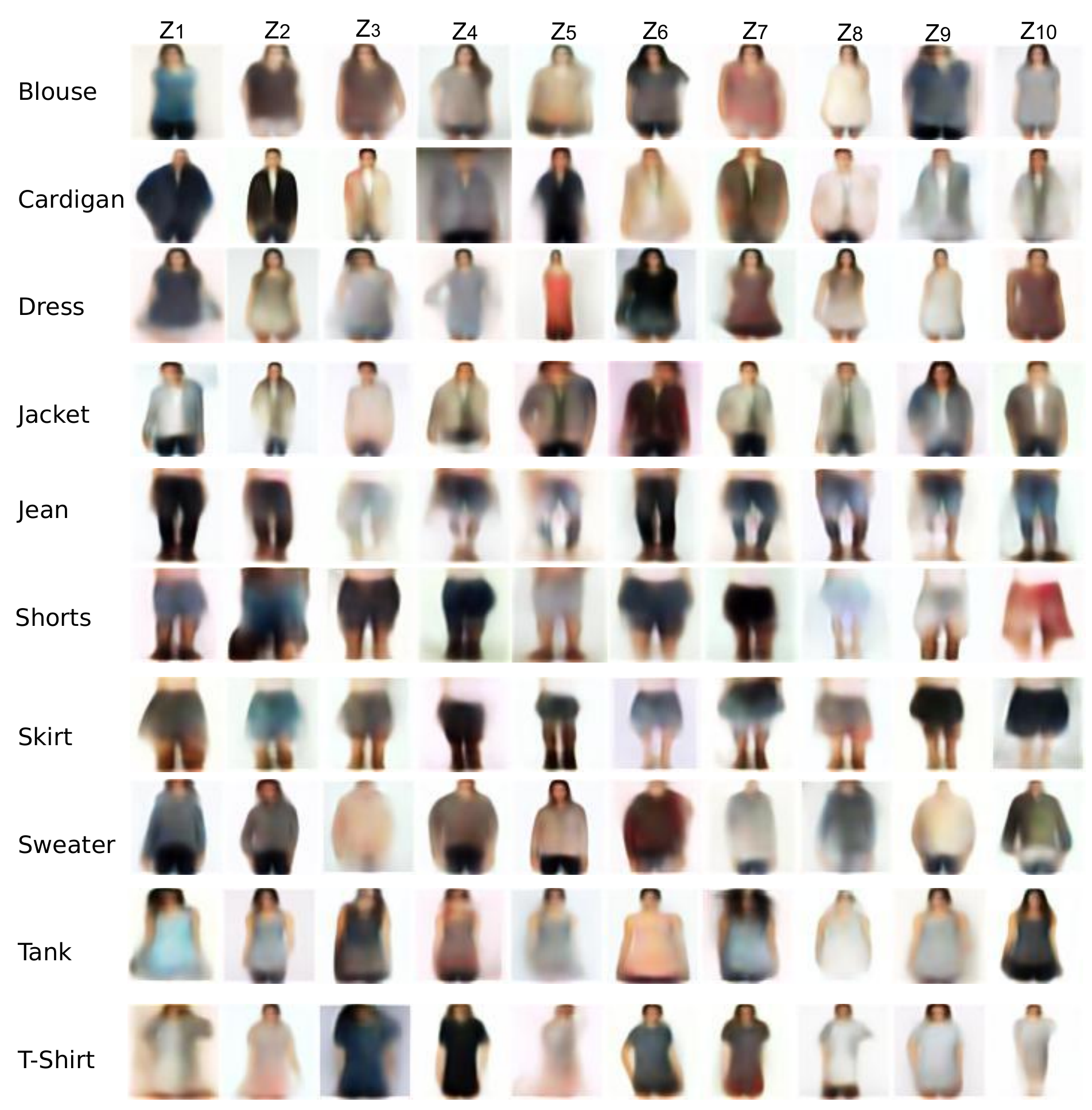}
\end{center}
   \caption{Each row is the decoded search target of a user for the given category using only top2 posteriors. Each column is for different samples of z from a normal distribution. As one can see the decoded search targets are distinctive from one another and they represent their corresponding categories properly.}
\label{fig:all}
\end{figure*}

\autoref{fig:ave}, \autoref{fig:short},~\autoref{fig:all} show qualitative results of our approach. 
As explained in~\autoref{sec:implementation}, noise is inherent in the gaze encoding and we therefore consider two pruning strategies to suppress minor activations.
Using directly the posterior of the CNN with gaze pooling layer, causes images that contains several categories rather than the intended visual search target.
As shown in~\autoref{fig:ave}, the image reconstructed from unaltered posteriors (first row) is more blurry and does not seem to contain one specific category (e.g. Z1 looks like a Blouse, Z6 is a dress and Z9 is a skirt). Images from top2 and top1 appear to be more focused on the intended category. Using top2 posteriors generates images which are a mixture between the two posteriors. One can see more details in top2 which is a composition between dress and skirt, whereas the top1 only contains skirt. 
Images from top1 are sharper and mostly contain one category. However, if the predicted category is wrong, we can not decode the intended category(last row of~\autoref{fig:short}). Also \cite{SattarBF16} showed strong recognition performance for top 1 and top 2 classification performance. Hence, using the top2 and top3, is likely to contain information about the target. 
This is reflected in better reconstructions for the top2 and top3 strategies.

As one can see in~\autoref{fig:short}, top1 decoded a dress, although the intended category was tank. The intended category is recovered in the top2 and top3 decodings. 

As top2 results are giving images with preferred search target, for further analysis, we chooses decoded images from top2. \autoref{fig:all} shows 10 samples for each category using top2 posteriors  based on fixations from one user. 
Our approach is able to generate images of different categories. The model performs better for several classes, as Jean, Shorts, Skirt, Dress, Tank, Sweater and Tee. Images from the cardigan and jacket are more similar to each other, although there are still differences in the appearance. In particular, images depicting the search target cardigan appear more elongated compared to those for jackets.  

\subsection{User Study: Search Target Recognition}
In order to assess the accuracy of the search target reconstruction we run a user study.
For each participant and category in the study we show 10 samples from our model. The reconstructions are based on human fixations from the dataset mentioned above. The users are asked to pick one category among 10 categories for a given image.  
The average accuracy of 19 users across categories was $62\%$ and the detailed confusion matrix is shown in~\autoref{tab:users2}. There are confusion between cardigan and jacket, also skirt and blouse with dress. Yet, all search targets were recognized significantly above chance (10\%). Users were most confident for jean, shorts and dress.

\begin{table}[h!]
\centering
 \begin{tabular}{c c c c c c c c } 
   &P1 & P2 & P3 & P4 & P5 & P6 & P7 \\ [0.5ex] 
 \hline
 Local &  \textbf{80\%}& \textbf{70\%} & \textbf{60\%} & \textbf{70\%}& \textbf{70\%}& \textbf{60\%} & \textbf{50\%} \\ 
 Global & 20\% &30\% & 40\% & 30\% & 30\% & 40\% & \textbf{50\%} \\ [1ex] 
\end{tabular}
 \caption{All of the users, preferred the decoding using local information over global method. This indicates the importance of local information on decoding the users' intents.}
 \label{tab:users1}
\end{table}

\begin{table}[h!]
\centering
\setlength{\tabcolsep}{4.5pt}
 \begin{tabular}{l | r r r r r r r r r r} 
   &Blouse& T-Shirt & Jean & Shorts & Skirt & \textcolor{blue}{Cardigan} & \textcolor{red}{Dress} & \textcolor{blue}{Jacket}&Sweater& Tanks \\ [0.5ex] 
    \hline
   \textcolor{red}{Blouse}& \textbf{42\%}& 5\%&0\%&0\%&5\%&0\%&\textcolor{red}{47\%}&0\%&0\%&0\% \\
   T-Shirt& 21\%& \textbf{74\%} & 5\%& 0\%&0\%&0\%&0\%&0\%&0\%&0\% \\
   Jean& 0\%& 0\% & \textbf{95\%}& 5\%&0\%&0\%&0\%&0\%&0\%&0\% \\
   Shorts& 0\%& 0\% & 0\%& \textbf{95\%}&5\%&0\%&0\%&0\%&0\%&0\% \\
   \textcolor{red}{Skirt}& 0\%& 0\% & 0\%& 0\%&\textbf{42\%}&0\%&\textcolor{red}{58\%}&0\%&0\%&0\% \\
  \textcolor{blue}{Cardigan} & 0\%& 0\% & 0\%& 0\%&0\%&\textbf{37}\%&0\%&\textcolor{blue}{58\%}&5\%&0\% \\
   Dress& 0\%& 16\% & 0\%& 0\%&5\%&5\%&\textbf{74\%}&0\%&0\%&0\% \\
   \textcolor{blue}{Jacket}& 0\%& 0\% & 0\%& 0\%&0\%&\textcolor{blue}{47\%}&0\%&\textbf{47\%}&0\%&5\% \\
   Sweater& 16\%& 0\% & 0\%& 0\%&0\%&10\%&0\%&16\%&\textbf{58\%}&0\% \\
   Tanks& 16\%& 0\% & 0\%& 0\%&5\%&5\%&21\%&0\%&0\%&\textbf{53\%} \\
\label{tab:users2}
\end{tabular}
\caption{Confusion Matrix of ``Search Target Recognition''. One can see in all of the cases, users were able to recognize the right categories above chance level $10\%$. (Bold number on diagonal corresponds to classification accuracy per class). However, Classes ``Blouse'' and ``Skirt'' are confused with ``Dress'' (in red). ``Jacket'' and ``Cardigan'' (in blue) where the other classes which users tend to be more confuse about them.}
\label{tab:users2}
\end{table}

\subsection{User Study: Local Vs Global Gaze Encoding}
In this user study,  we test the importance of local gaze information for the decoding of visual search targets. Our full model is denoted as ``local'' here, as it uses the full gaze information -- in particular the fixation location on the image. We compare it to a ``global'' model which uses gaze information -- but only to the extend that we know an image was fixated without knowing the exact location. This also connects to the analysis performed on the recognition task for the gaze pooling layer in \cite{SattarBF16}. We ask how much of a difference these approaches make in terms of search target reconstruction.

\begin{figure*}
\begin{center}
\includegraphics[width=1\linewidth]{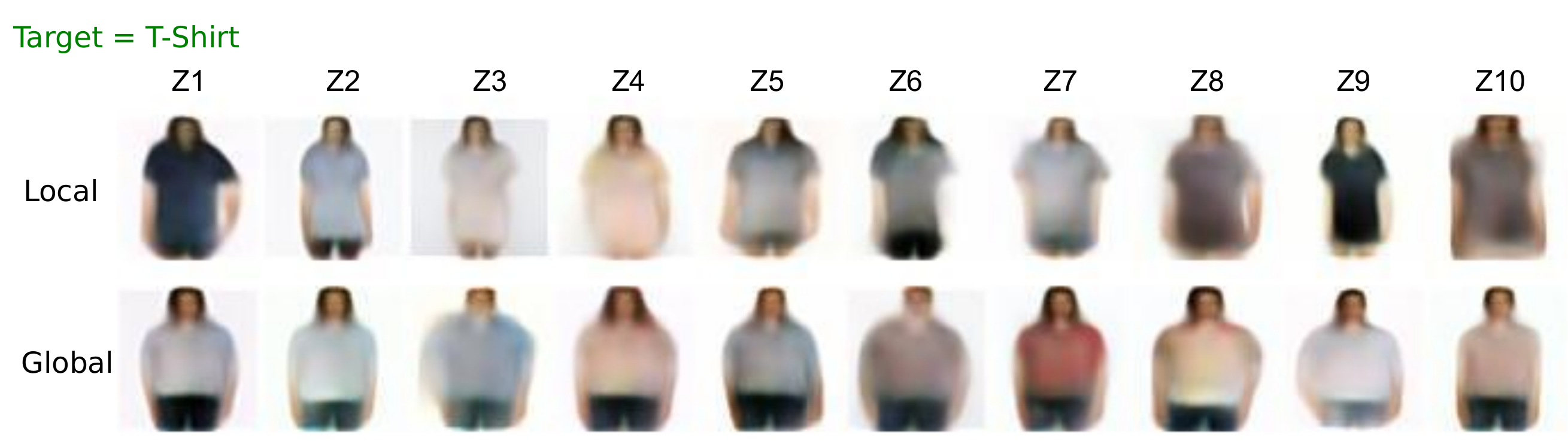}
\end{center}
   \caption{Example image used in our second user study. For each category, users need to select between local and global decoded target. Local method encodes the gaze data using gaze-pooling layer which benefits from user intended local image regions.}
\label{fig:exp}
\end{figure*}

In this user study, each participant saw two rows of search target reconstructions.  One row was generate by the local, the other by the global method (~\autoref{fig:exp}.
The  users were  instructed to select the row which matches the best, the given search target category. 
The users selected the local encoding method in $65\%$ of the cases. The chance level for this experiment is $50\%$ as for each image the participates do a binary task. Consequently, the gain is $65\%/50\% =13\%$. The gain indicates how much performance of users differs from a random selections. Also, we performed Chi-Square Goodness of Fit Test, to investigate the significant of our result. The null hypothesis was that both local and global decoding are equal. The $\chi^2$ value is 6.914 and P-Value is 0.009. Hence, the result is significant at $p\le 0.05$ and therefore local information is key for improved search target reconstruction. Detailed results are shown in ~\autoref{tab:users1}.

\section{Conclusion}
In this paper, we introduce the first approach to decode visual search target of users from their gaze data. This task is very challenging as the target only resides in user mind. For this aim we used recent advances in generative image models and search target prediction using gaze data.  Proof of concept is established by two user studies, showing that the decoded target lead to human recognizable visual representations as well as highlighting the importance of localized gaze information. We like to emphasize that due to the training setup, the method remains highly practical and applicable, as no large scale gaze data had to be collected or used. Key is rather the utilization of a semantic layer that connects the gaze encoder with the conditional generative image model.